\documentclass[11pt,a4paper]{article} 
\usepackage{geometry}
\usepackage{graphicx}
\usepackage{parskip}
\usepackage{textcomp}
\usepackage{color}   
\usepackage[utf8]{inputenc}
\usepackage{authblk}
\usepackage{amsmath}
\usepackage{color}
\usepackage{caption}
\date{}

\title{Word-length entropies and correlations \\of natural language written texts}

\author[1]{Maria Kalimeri\thanks{corresponding author, now at Laboratoire de Biochimie Th\'{e}orique, IBPC, CNRS, UPR9080,  Univ. Paris Diderot, Sorbonne Paris Cit\'{e}, France, e-mail: maria.kalimeri@ibpc.fr}}
\author[2]{Vassilios Constantoudis}
\author[3,4]{Constantinos Papadimitriou}
\author[4]{Konstantinos Karamanos}
\author[4]{Fotis K. Diakonos}
\author[5]{Harris Papageorgiou}
\affil[1]{Department of Mathematics, National Technical University of Athens, Greece}
\affil[2]{Department of Microelectronics, NCSR Demokritos, Aghia Paraskevi, Greece}
\affil[3]{IAASARS, National Observatory of Athens, Penteli, Greece}
\affil[4]{Department of Physics, University of Athens, Greece}
\affil[5]{Institute for Language and Speech Processing, Athena R.C., Greece}

\begin{document}
\maketitle

\begin{abstract}
We study the frequency distributions and correlations of the word lengths of ten European languages. Our findings indicate that a) the word-length distribution of short words quantified by the mean value and the entropy distinguishes the Uralic (Finnish) corpus from the others, b) the tails at long words, manifested in the high-order moments of the distributions, differentiate the Germanic languages (except for English) from the Romanic languages and Greek and c) the correlations between nearby word lengths measured by the comparison of the real entropies with those of the shuffled texts are found to be smaller in the case of Germanic and Finnish languages. 
\end{abstract}

% \smallskip
% \noindent \textbf{Keywords:} word-length representation, language family, n-gram entropies, word-length distributions, word-length correlations

\section{Introduction}
In quantitative linguistics, the word length has been considered from the very beginning a key feature whose systematic study can reveal important aspects of language structure, differentiations and text typology. Since 1851, when August de Morgan proposed the study of word lengths as a hallmark of the text style and a possible factor in determining authorship, a great amount of relevant work has been undertaken and published (for a nice review see \cite{GrzybekA180}). The principal aim of these studies has been to introduce and validate theoretical models providing a sufficient description of the frequency distribution of the word lengths of various samples (ordinary or frequency dictionaries, texts of different types). These models can then be used for revealing inter-language or intra-language trends and affinities  \cite{mikros, ktori, eger, grzybek, pande,rottmann, riedemann,ziegler}.

During the last two decades, word-length representations of written texts have also been investigated in the context of recent advances in time-series analysis and the theory of complex systems. The relevant few studies are differentiated from the previous linguistic ones in two ways. The first is the unit of word-length measurement.  Contrary to the main trend in quantitative linguistics of measuring the word syllables, these studies count the number of word letters (graphemes). Alternatively, the recorded variable in the position of a word can be the rank according to its frequency of appearance \cite{mont2002,zanette05,mont2011,ausloos2007, ausloos2008, ausloos2012} or the sum of the Unicode values of the letters of a word \cite{sahin}. Secondly, the studies of this category raise the issue of word correlations along the text and put emphasis on the detection and quantification of the long-range correlations in the word ordering of a text. The methodology of these studies involves mainly Hurst and detrended fluctuation analysis for the detection and characterization of fractal and multifractal properties along with measurements of $n$-gram entropies. 

In this paper, we are trying to bring together these approaches combining selectively their methods with the aim to investigate the effects of text language on the metrics of word-length distributions and correlations. The texts are taken from the Europarl corpus comprising European Parliament parallelized texts of the same content written in ten different European languages belonging to the Romanic (French, Italian, Spanish, Portuguese), Germanic (English, Dutch, German, Swedish), Uralic (Finnish) and Greek languages \cite{koehn2005}. 
More specifically, we take into consideration  the frequency distributions of word lengths and calculate their  moments and entropies to quantify their characteristics. Furthermore, inspired by the well-known Zipf diagram we plot the frequency of word lengths versus their rank and focus on the behavior of the high ranks (low frequencies). These Zipf-like diagrams will help us explain the observed trends in distribution moments and entropy. 

Following the complex-system approach, we estimate the word length by measuring the letters of the words and calculate the $n$-gram entropies of the word-length series emanating from the sample texts. In addition, we compare their values with those of the shuffled data to quantify the correlations between nearby word lengths. The emphasis is put on the short-range word-length correlations (herein bigrams and trigrams) complementary to the recent work by M.A. Montemurro and D.H. Zanette \cite{mont2011} where a similar analysis has been performed focused on the long-range correlations.  

A first application of the bunch of these methods to the statistical and correlation analysis of texts in two languages (English and Greek) and different genres has been presented in a previous work of our group \cite{kalimeri2012}. In the present work, we extend our study to a much richer corpus comprising parallel texts in ten languages and exploit the content similarity to isolate the effects of language family on word-length distributions and correlations. 

The paper is organized as follows: In the next section we describe the corpus to be analyzed and in section 3 we report the methods we are using. The results are presented in section 4 grouped in two subsections. The first, includes the results for the word length distributions and their metrics (moments and unigram entropies) along with the respective Zipf-like diagrams. The second subsection concerns the results of bigram and trigram entropies and the comparison with the shuffled texts in order to estimate the nearby word-length correlations. The paper closes with a summary of the main findings in section 5.

\section{Corpus description}
For our calculations, we use part of the Europarl parallel corpus, which was extracted from the proceedings of the European Parliament \cite{koehn2005}. The whole corpus includes versions in 21 European languages: Romanic (French, Italian, Spanish, Portuguese, Romanian), Germanic (English, Dutch, German, Danish, Swedish), Slavic (Bulgarian, Czech, Polish, Slovak, Slovene), Uralic (Finnish, Hungarian, Estonian), Baltic (Latvian, Lithuanian), and Greek. Europarl is offered with bilingual alignments of all languages to English. Herein, we analyze parallel texts (translated versions of the same text) of ten Europarl languages (see Table \ref{corpora}); our assumption is that possible spotted differences are due to language characteristics and not due to meaning and semantics.

\begin{table}[ht]
\caption{Breakdown of the corpus \label{corpora}}
\begin{center}
  \begin{tabular}{  l  l }
    \hline \\ [-1.5ex]
Language & \# of words  \\ 
    \hline 
    \hline \\ [-1.5ex]
Finnish (fi)     & 10.15M \\
German (de)  &  14.68M \\
Swedish (sv)  &   14.23M\\
Dutch (nl)      &  16.23M\\
English (en)   &  15.98M\\
Italian (it)      &  15.57M\\
French (fr)    &   17.55M\\
Spanish (es)  &   16.22M \\
Portuguese (pt)  & 16.06M \\
Greek (el)     & 15.85M \\
\hline
Total & 152.52M \\
  \end{tabular}
  \label{table1}
\end{center}
\end{table}

\section{Methodology}
Our methodology consists of the following steps:
\subsection{Conversion of texts into word-length series}
\begin{figure}[h]
\centerline{\includegraphics[width=.6\textwidth]{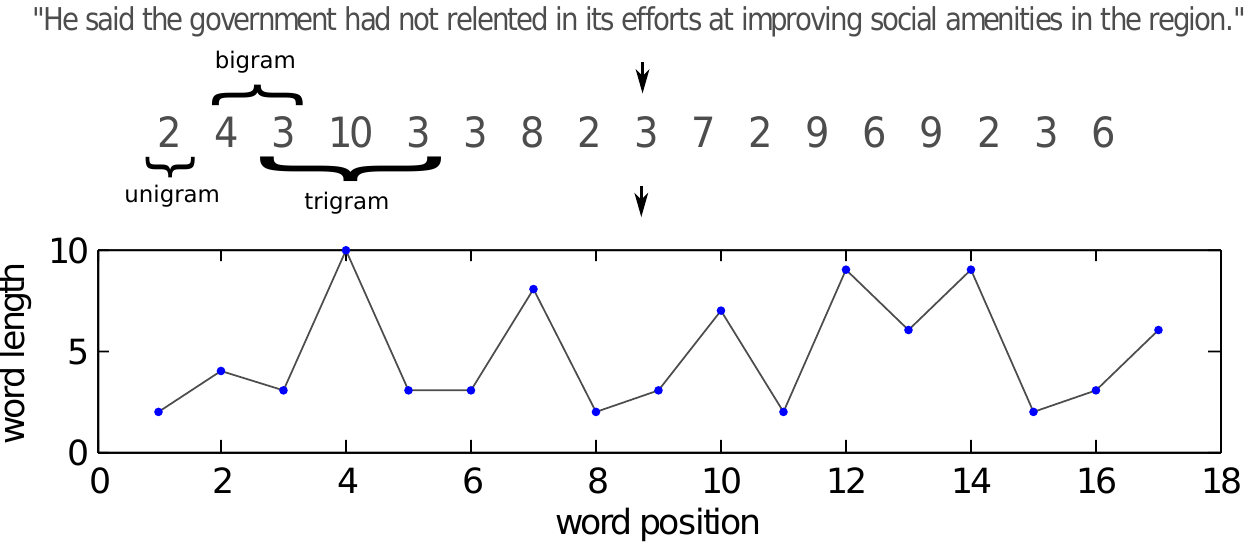}}
\caption{Figure 1: An example of the mapping of a text to a word-length series along with the definition of unigram, bigram and trigram. \label{schema}}
\end{figure}

We construct the time series from the corpus data by mapping each document to a sequence of numbers $w_i$, $i=1,\ldots,N$ where every number represents the length of the respective word. The resulting sequence consists of integers with a minimum equal to 1 and a maximum equal to the length of the longest word in the specific language corpus. An example of this representation is shown in Fig.  \ref{schema}.

The word-length series are then studied both in terms of their frequency distributions and their use in the estimation of the $n$-gram entropies as defined below. Let us note, that due to the nature of the corpora, the resulting sequences for each language (discrete time series) incorporate the speaker diversity and topic variety of the source corpus.

\subsection{Word-length frequency distributions}
After obtaining the word-length series of all languages, we estimate the normalized frequency (i.e. probability) of each word length and we plot the word-length frequency distribution for each language. Then, we calculate the first four moments of the distributions as well as their entropy.
The first moment coincides with the mean value of the word lengths $\langle w_i^l \rangle$, where $\langle \ldots \rangle$ indicates averaging over the entire word-length series of the specific language $l$. The second moment is the square of the standard deviation (variance), $(sd^l)^2 = \langle (w_i^l)^2 \rangle -\langle w_i^l \rangle ^2$, and quantifies the width of the frequency distribution. The third ($n=3$) and fourth ($n=4$) moments, $m_n^l= \langle (w_i^l - m_1^l)^n\rangle$, are related to the skewness, $sk^l= m_3^l/(sd^l)^3$, and kurtosis, $kurt^l= m_4^l / (sd^l)^4$, respectively. The skewness characterizes the asymmetry of the distribution, being zero at perfectly symmetric distributions. Large positive (negative) values of skewness characterize distributions with long tails at the right (left) of the mean value. The kurtosis is related to the peakedness and tailedness of the distribution and large values reveal distributions with high, steep peaks and long, thick tails. 

We now return to our word-length series. In order to retain the short-discussion character of the Europarl corpora, we go further into fragmentizing the series in segments of N = 1000 words. Besides, in the framework of language identification, the available texts are usually much smaller than our source corpora. The moments of the distributions are then calculated over each one of these segments and the final value results from averaging over all segments of the same language.  
    
\subsection{Zipf-like diagrams of $n$-gram word-length frequencies}
In the spirit of the Zipf’s diagram, where the frequency of a word in a corpus is plotted against its rank in the frequency table, we investigate similar diagrams with the word-length frequencies and generalize the same concept in the case of word-length $n$-grams (see also Fig.  \ref{schema} for $n$-gram definition). In particular, we first sort the $n$-grams appearing in the word-length series of each language in a frequency list. Then we plot the frequencies of the $n$-grams with respect to their rank in the corresponding list.  Here, we limit our analysis to unigrams, bigrams and trigrams ($n=1,2,3$). The plots are presented in semilogarithmic scales to distinguish between frequencies of word lengths with high ranks.

\subsection{$N$-gram entropies}
The $n$-gram entropies are block entropies that extend Shannon’s classical definition of the entropy of a single state (unigram) to the entropy of a succession of states ($n$-grams) \cite{nicolis1994, karamanos1999}. Actually, they account for the form of the probability distribution of $n$-grams and large entropy values imply more uniform distributions of the appearance probabilities of $n$-grams.

In the general case, following the ``gliding" reading procedure, the calculation of $n$-gram entropies is done as follows \cite{nicolis1991,nicolis2005,nicolis2007}. For the word-length series $w_i^l$, $i=1,\ldots,N_l$ we set $K = N - n + 1$ and define the $n$-grams $n_j = (w_j, \ldots , w _{j+n-1})$ where $j = 1, 2, \ldots ,K$. The (Shannon-like) $n$-gram entropy is defined as\\
\begin{equation}
\Phi_n=-\sum_{(s_1,...,s_n)}p^{(n)}(s_1,...,s_n)\ln(p^{(n)}(s_1,...s_n))\,\,
.\label{this}
\end{equation}
where the summation is over all unique $n$-grams and the probability of occurrence of a unique $n$-gram ($w_1, . . . , w_n$), denoted by $p^{(n)}(w_1, . . . , w_n)$ is defined by the fraction
\begin{center}
$\frac{\text{No. of $n$-grams }(s_1,\ldots,s_n)  \text{ encountered when gliding}}{\text{Total No. of $n$-grams}}$
\end{center}
This quantity is maximized for the so-called {\it normal sequence} where all possible $n$-grams appear with equal probability \cite{karamanos2005,borwein2008}.

Just as for the moments of the distributions, the block entropies are also calculated over each one of the 1000-word segments and the final value results from averaging over all segments of the same language. 

\subsection{Short-range word-length correlations $C_n$}
$N$-gram entropies for $n>1$ (here $\Phi_2$ and $\Phi_3$) contain information about the word-length ordering in texts, but are also strongly affected by the unigram (single word length) frequencies. In order to isolate correlation effects, one common procedure is to generate a new word-length series through shuffling of the words of the original series i.e. to interchange the word positions  in a random and uncorrelated manner and then compare the $n$-gram entropy of the new series with that of the original one. By definition, the shuffling process will give more uniform $n$-gram distributions and therefore larger entropies. The difference of the $n$-gram entropy of the shuffled series and the entropy of the original series, $C_n = \Phi_{n,shuffled} - \Phi_n$, can be considered a measure of the correlations between  the word-lengths inside the $n$-gram.  A similar quantity, the relative entropy, has been estimated by subtracting the entropy rate per word of the original texts from that of the shuffled texts in \cite{mont2011} and used as a metric of the word ordering correlations at scales larger than the sentence. In the present work, we calculate the differences $C_2$ and $C_3$ of the bigram and trigram entropies to quantify the correlations between nearby word lengths in the analyzed corpus. Therefore, the focus here is on the word ordering patterns within sentences. 

\section{Results}
\subsection{Word-length frequency distributions and metrics}
\begin{figure}[h]
\centerline{\includegraphics[width=1.0\textwidth]{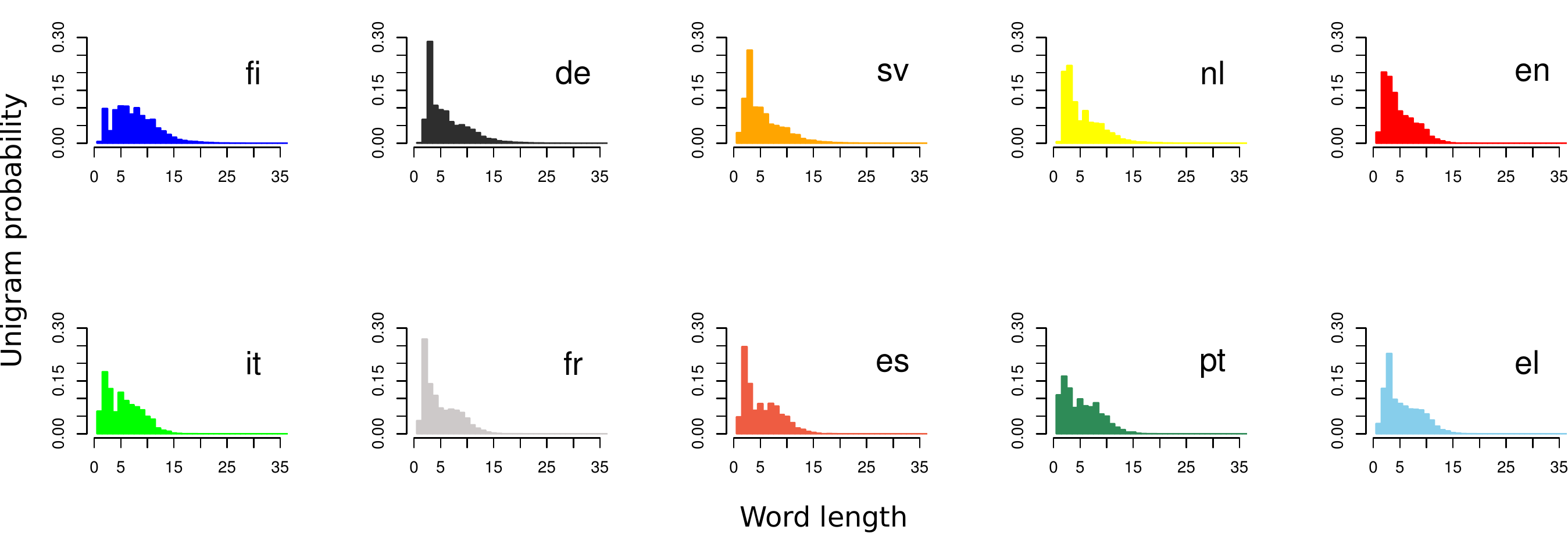}}
\caption{Figure 2: The word-length distributions of the parallel texts from Europarl corpus in ten languages: Finnish (fi), German (de), Swedish (sv), Dutch (nl), English (en), Italian (it), French (fr), Spanish (es), Portuguese (pt) and Greek (el). Unusually large words found in all languages correspond to chemical compounds. \label{uniDist}}
\end{figure}

The probability distributions of the word lengths obtained from the corpora for the ten languages of the Europarl corpus are shown in Fig. \ref{uniDist}. In almost all languages, a pronounced  peak at short word lengths (2-3 letters) is observed with the striking exception of the Finnish distribution where there is no peak but a dip at word length 3 and a quite wide plateau at the middle word lengths from 4 to 8 letters. The height of the peak is smaller in Portuguese and Italian and larger in French, German and Swedish distributions. It is worth noting, that in case of Greek, our results are in compliance with the work in \cite{mikros}, one of the few approaches using the number of letters as word-length unit.

\begin{figure}[ht]
\centerline{\includegraphics[width=0.8\textwidth]{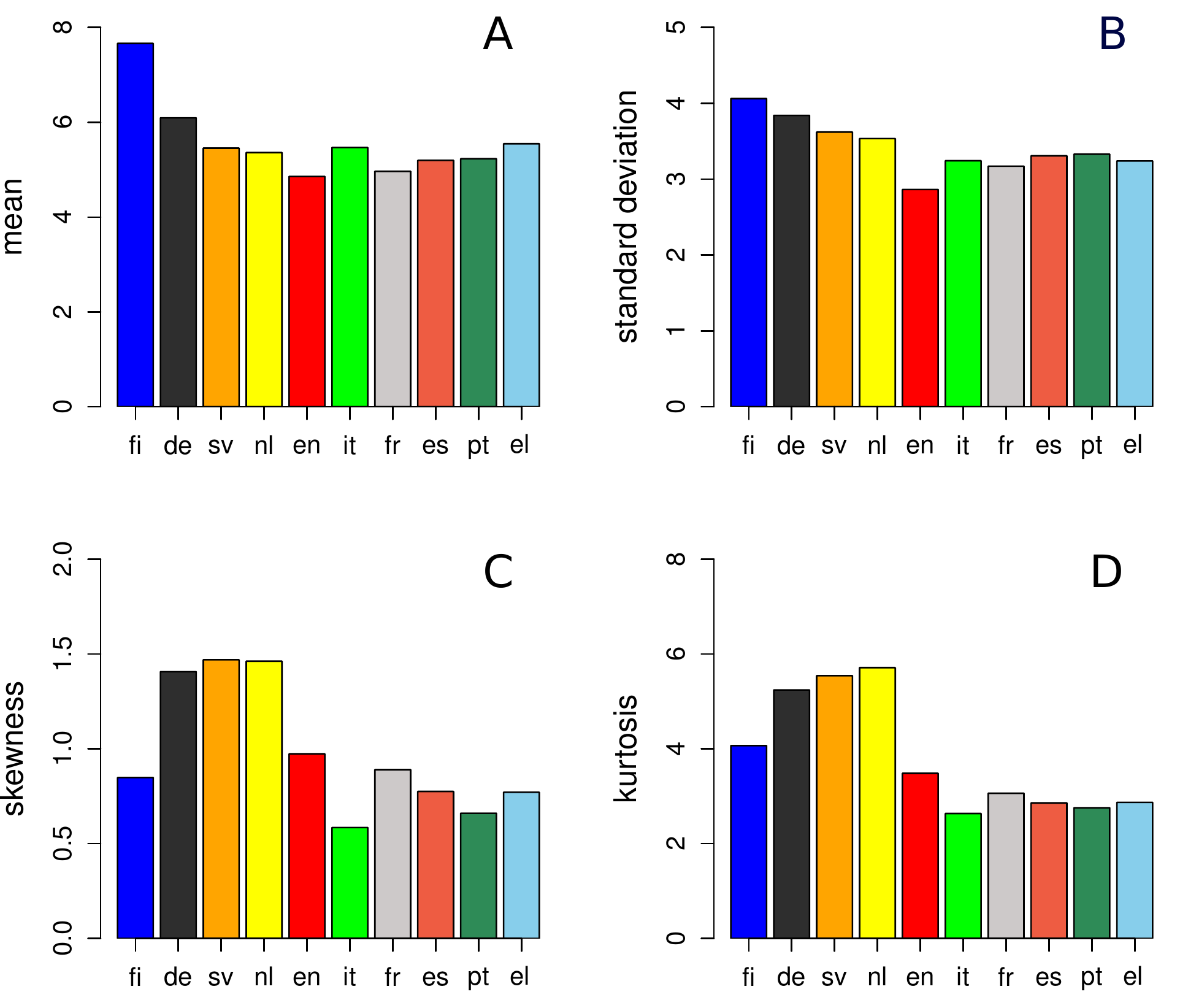}}
\caption{Figure 3: The mean word length (A) and the standard deviation (B), the skewness (C) and the kurtosis (D) of the word-length frequency distributions for the ten languages of the Europarl corpus. \label{moments}}
\end{figure}

The mean word lengths of all languages are shown in Fig. \ref{moments}A. Not surprisingly, the Finnish corpus possesses the highest mean value ($\sim8$) due to the agglutinative character of the Finnish language followed by the German texts with mean word length slightly larger than 6 letters. The smallest mean is found in the English corpus (4.9), while the rest of the languages have means fluctuating between 5 and 6 letters. With the exception of the Finnish language and possibly all Uralic languages, no other clear sign of the language family has been detected in the mean word length since Greek, Italian, Swedish and Dutch have very similar mean word lengths despite the different family they belong to. The situation changes in the metrics related to higher order moments: standard deviation (Fig. \ref{moments}B), skewness (Fig. \ref{moments}C) and kurtosis (Fig. \ref{moments}D). Gradually, as we move to higher order moments, from standard deviation to kurtosis, the Germanic languages (except for English) take clearly larger values with respect to Romanic and Greek languages. This means that the frequency distributions of the Germanic languages are wider and less symmetric than those of Romanian and Greek languages. In addition, the differences in kurtosis show that the distributions of Germanic languages are platykurtic (kurtosis$>3$) whereas the Romanic and Greek languages are slightly leptokurtic (kurtosis$\leq3$). 

Quite interestingly, the English skewness and kurtosis lie in between the values of the other Germanic languages (German, Swedish and Dutch) and the corresponding values of the Romanic and Greek languages, revealing this way the strong influence of the latter on its lexicon. The Finnish moments exhibit a peculiar behavior with respect to the other languages mainly coming from the absence of peak and the presence of the plateau in the word-length probability distribution. Thus, the standard deviation gets the highest value similarly to the mean word length, whereas the skewness and kurtosis lie almost in the middle of the total range spanning from the values of all corpora.     

In order to investigate the effect of the block length on our findings, we have calculated the values of the moments without the segmentation and found them differing less than 1$\%$ from those with the 1000-word segmentation. Additionally, we plot the distributions of each moment over all segments of length 1000, for each language in Figure S1 in the supplementary material.

The high order moments of a distribution are dominantly defined by the characteristics of its peak and tails. Given that the peak in the distributions of all languages (except Finnish) lies at the word length 2-3 with no clear  effect of language family, the observed differences can only be attributed to the behavior of the tail at long word lengths. 

\begin{figure}[ht]
\centerline{\includegraphics[width=.6\textwidth]{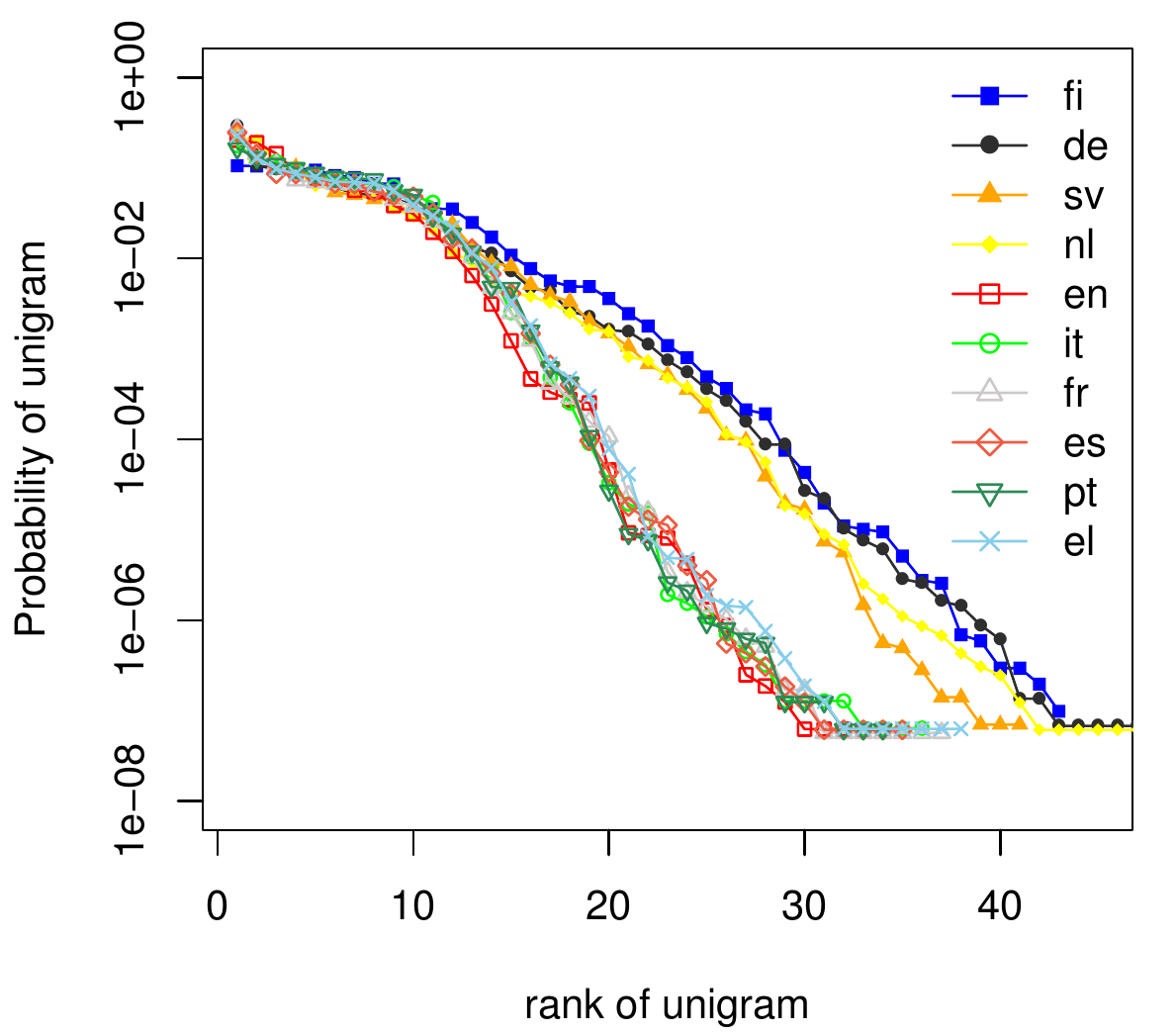}}
\caption{Figure 4: Zipf-like diagram of the unigram (single word length) probability versus its rank. \label{rankUni}}
\end{figure}

In order to check this hypothesis, we plot and investigate the Zipf-like diagrams of the normalized frequencies of the word lengths against their rank and focus on the behavior at high ranks. The results are shown in Fig. \ref{rankUni}. One can clearly notice the grouping of the frequency curves at high ranks in two groups: The first includes the Romanic languages along with the Greek and English and exhibit smaller probability of long words with respect to the second group comprising the Germanic and the Finnish languages. Therefore, it seems that the more frequent appearance of rare long words in the Germanic languages have a clear and pronounced effect on the high-order moments of their word-length distributions resulting in higher values of the skewness and kurtosis. 

Finally, we calculate the unigram entropies $\Phi_1$ of all distributions. The unigram entropy $\Phi_1$ characterizes the uniformity of the distributions shown in Fig. \ref{uniDist} and therefore we expect that Finnish would exhibit the highest $\Phi_1$ followed by the Portuguese and Italian. The expectation is verified by the numerical results shown in Fig. \ref{wordLengthSeries} (see also the left panel of Figure S2 of the supplementary material depicting the distribution of the block entropies over all the different segments).

Despite the sensitivity of the unigram entropy to the text language, only Finnish and possible other Uralic languages can be distinguished in terms of mean word-length comparisons. Given the fact that the entropy is mainly determined by the unigrams of high frequencies (short word lengths), we draw the conclusion that we cannot distinguish Germanic from Romanic languages on the basis of the short word-length distributions.

\begin{figure}[ht]
\centerline{\includegraphics[width=.6\textwidth]{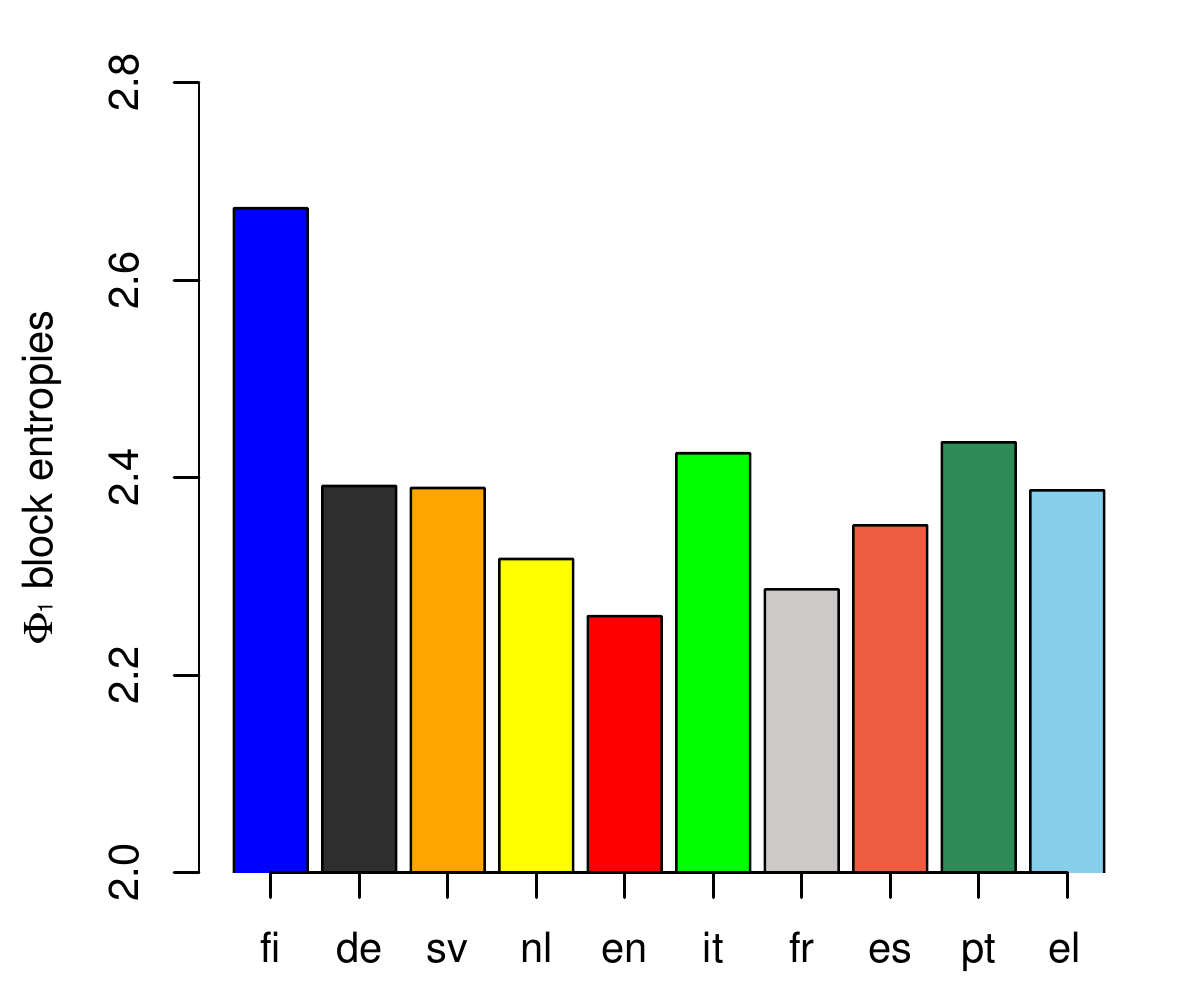}}
\caption{Figure 5: Unigram Shannon entropies of the analyzed languages from the Europarl corpus. \label{wordLengthSeries}}
\end{figure}

In summary, our results so far have shown that the short/middle and large word-length parts of the frequency distributions are sensitive to different language families. For example, the Finnish corpus can be easily distinguished when we are looking at the short/middle word-lengths and the parameters defined by these such as the mean word-length and the entropy. On the other hand, the tails of the frequency distribution with the rare large word-lengths seem to be able to separate the Germanic from the Romanic, Greek and Finnish languages. 

\subsection{$N$-gram entropies and Zipf-like diagrams for $n=2, 3$}
\begin{figure}[ht]
\centerline{\includegraphics[width=.9\textwidth]{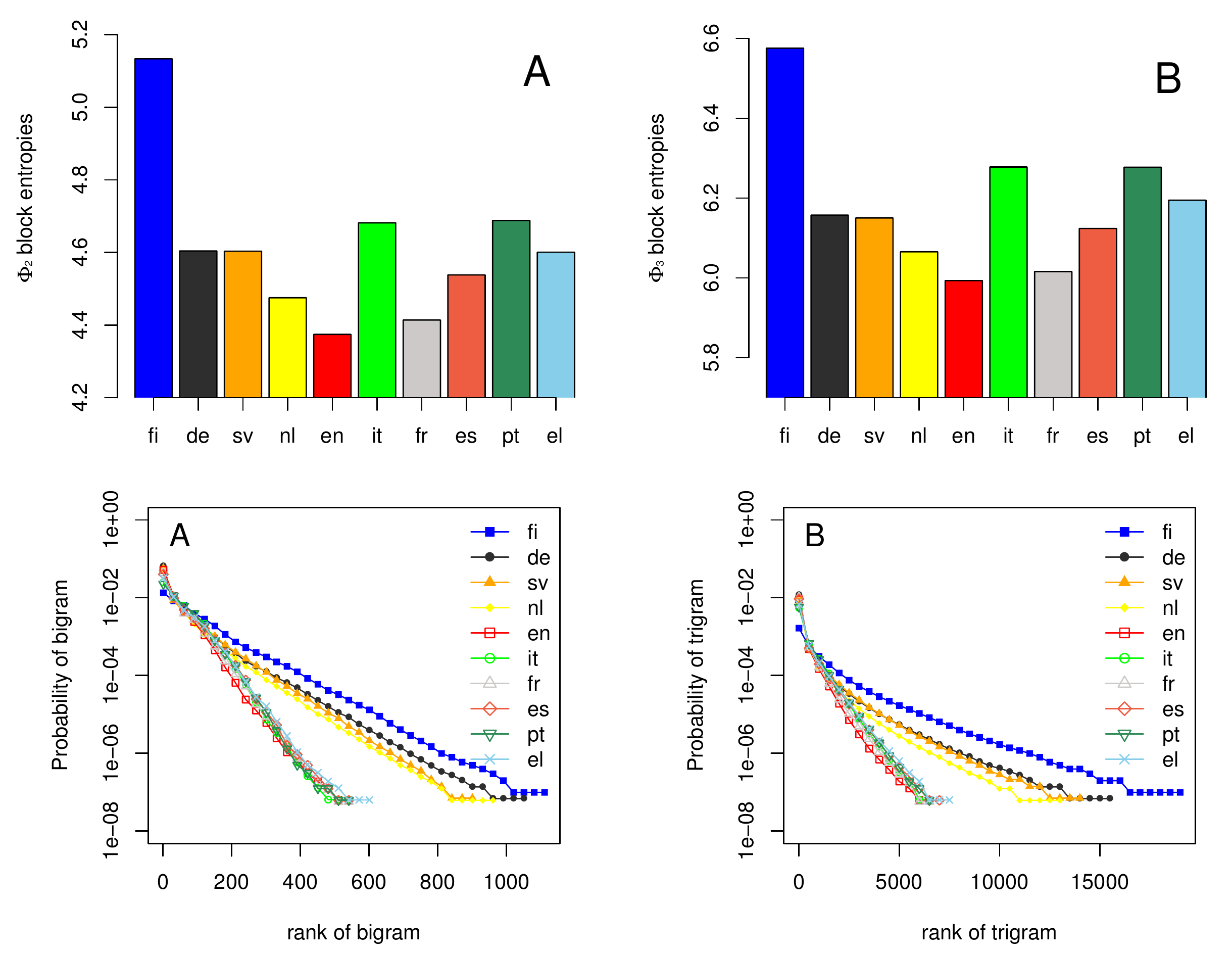}}
\caption{Figure 6: Top panel: Bigram $\Phi_2$ (A) and trigram $\Phi_3$ (B) entropies of the analyzed languages from the Europarl corpus. Lower panel: Zipf-like diagrams of the frequency (probability) of bigrams (A) and trigrams (B) with respect to their frequency rank.  Notice the grouping of the curves at high ranks according to the family of the language (Finnish, Romanian, German, Uralic).\label{bitri}}
\end{figure}

Before going to the correlations between the positions of word-lengths in the corpora, we calculate the bigram ($n=2$) and trigram ($n=3$) entropy for all languages along with the Zipf-like diagrams of the frequency distributions of bigrams and trigrams. As explained in the previous section, $n$-gram entropies, where $n>1$, incorporate the effects of both word-length frequencies and correlations. Figure \ref{bitri}, at the top panel, shows the values of the bigram ($\Phi_2$) and trigram ($\Phi_3$) entropies for all languages (the $\Phi_2$ and $\Phi_3$ distributions are shown in the central and right-most panels of Figure S2 in the supplementary material). One can clearly notice that both entropies follow the same language ordering to unigram entropy $\Phi_1$. This means that word-length frequencies dominate in the language dependence of the $n$-gram entropies at least for $n=2$ and $n=3$. The impact of correlations is weaker, but, as will be indicated in the next section, reveals important sensitivity to text language.

The Zipf-like diagrams of the frequency distributions of bigrams and trigrams in semi-log axes are shown in the lower panel of Fig. \ref{bitri}. Again we get a similar picture to the unigram Zipf-like plot of Figure \ref{rankUni}. At high ranks (large word lengths) the curves are grouped according to their family (Uralic/Finnish, Germanic and Romanic/Greek). The only exception is the English for reasons explained above (see 4.1).

\subsection{Word-length correlations}
\begin{figure}[ht]
\centerline{\includegraphics[width=\textwidth]{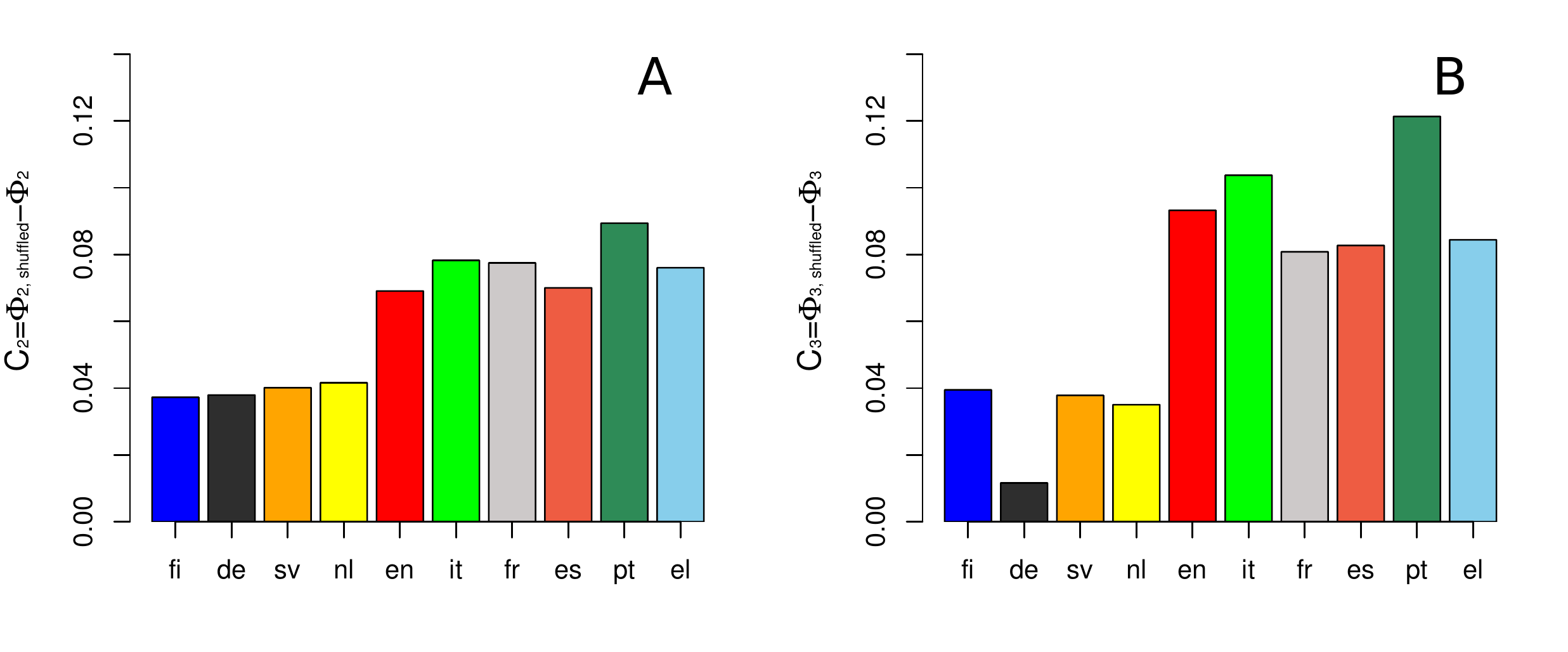}}
\caption{Figure 7: The differences $C_2$ and $C_3$ of the entropies of the source corpora, $\Phi_2$ and $\Phi_3$, from the respective entropies of the shuffled corpora $\Phi_{2,shuffled}$ and $\Phi_{3,shuffled}$ for all analyzed languages. $C_2$ and $C_3$ quantify the short-range correlations between nearby word lengths. Notice the smaller $C_2$ and $C_3$ values (weaker calculations) of the Germanic and Finnish languages with respect to the Romanian, Greek and English languages.  \label{c2c3}}
\end{figure}

To isolate the short-range correlations in the word-lengths series of the different languages and quantify them, we calculate the $C_2$ and $C_3$ quantities by subtracting the original $\Phi_2$ and $\Phi_3$ values from the $\Phi_2$ and $\Phi_3$ of the shuffled texts. The results are shown in Fig. \ref{c2c3} and indicate that the Germanic languages along with the Finnish have systematically smaller $C_2$ and $C_3$ i.e. they exhibit weaker correlations between nearby word lengths from the Romanian, English and Greek language. Similarly to the high-rank behavior of frequency curves reported in section 4.2, the English $C_2$ and $C_3$ are closer to the Romanian and Greek languages revealing the affinity of English to Romanic languages even in the nearby word-length correlations.  

The results of Fig. \ref{c2c3} indicate that at the very short scale of two to three nearby words, the word patterns do not exhibit the universality proposed by Montemurro and Zanette, which they detect at longer, extra-sentence scales \cite{mont2011, montemurro2013}. On the contrary, they demonstrate the influence of language family supporting at the same time the significance of cultural effects on nearby word ordering patterns \cite{dunnevolved2011}.

\section{Conclusions}
In this work we study the effects of language family on the distributions and correlations of word lengths. In particular we analyze the word length representation of a large parallel Europarliament corpus translated in 10 different European languages, belonging to the Romanic (French, Italian, Spanish, Portuguese), Germanic (English, Dutch, German, Swedish), Uralic (Finnish) and Greek languages. We find that, the Uralic (Finnish) corpus can be distinguished on the bases of short and more frequent words which dominate the mean word length and the values of the block entropies. In contrast to that, due to the large and more rare word lengths, the family of Germanic languages, with the notable exception of English, stand out when we focus on higher order moments of the distributions, namely standard deviation, skewness and kurtosis. Finally, the short-range word-length correlations, measured by the difference of the actual block entropies from those of the shuffled versions of the texts, are found to be smaller for both Germanic and Finnish languages. 

Over all we see that, depending on the metric, the word-length representation is capable not only to distinguish among language families but also to detect deviations originating in cultural influences, such as the impact of Romanic and Greek languages on the English lexicon. On the contrary, a different symbolic representation such as the commonly used frequency-rank series, might fail to produce as fruitful results for small corpora in combination with either distribution moments (due to the very long tails and the power law nature of the distributions) or bigram and trigram entropies (due to the huge number of possible bigram and trigram realizations).

Future work may be oriented towards two directions. In the first, we can elaborate more on the found differences between languages and investigate their contribution to the problem of automatic language identification. In the second direction, one can focus on the issue of word-length correlations at short (intra-sentence) and long scales via the calculation of the relative entropy and correlation metric $C_n$ for $n>3$ to enable a more profound comparison with the findings presented by Montemurro and Zanette in \cite{mont2011, montemurro2013}.  \\

{\large \bf Acknowledgements}

We are grateful to George Mikros for the fruitful discussions and support. K.K. would like to express his gratitude to the Library in the National Hellenic Research Foundation.\\

\bibliographystyle{ieeetr}

\begin{thebibliography}{10}

\bibitem{GrzybekA180}
P.~Grzybek, ``History and methodology of word length studies. {T}he state of
  the art,'' in {\em Contributions to the Science of Text and Language. Word
  Length Studies and Related Issues}, vol.~31 of {\em Text, Speech and Language
  Technology}, pp.~15--90, Dordrecht, NL: Springer, 2006.

\bibitem{mikros}
G.~K. Mikros, N.~Hatzigeorgiu, and G.~Carayannis, ``Basic quantitative
  characteristics of the modern greek language using the hellenic national
  corpus,'' {\em Journal of Quantitative Linguistics}, vol.~12, no.~2-3,
  pp.~167--184, 2005.

\bibitem{ktori}
M.~Ktori, W.~Van~Heuven, and N.~Pitchford, ``Greeklex: A lexical database of
  modern greek,'' {\em Behavior Research Methods}, vol.~40, no.~3,
  pp.~773--783, 2008.

\bibitem{eger}
S.~Eger, ``A contribution to the theory of word length distribution based on a
  stochastic word length distribution model.,'' {\em Journal of Quantitative
  Linguistics}, vol.~20, no.~3, pp.~252--265, 2013.

\bibitem{grzybek}
P.~Grzybek and E.~Stadlober, ``Project report: The graz project on word length
  (frequencies),'' {\em Journal of Quantitative Linguistics}, vol.~9, no.~2,
  pp.~187--192, 2002.

\bibitem{pande}
H.~Pande and H.~S. Dhami, ``Model generation for word length frequencies in
  texts with the application of zipf's order approach.,'' {\em Journal of
  Quantitative Linguistics}, vol.~19, no.~4, pp.~249--261, 2012.

\bibitem{rottmann}
O.~A. Rottmann, ``Word and syllable lengths in east slavonic,'' {\em Journal of
  Quantitative Linguistics}, vol.~6, no.~3, pp.~235--238, 1999.

\bibitem{riedemann}
H.~Riedemann, ``Word-lengt distribution in english press texts,'' {\em Journal
  of Quantitative Linguistics}, vol.~3, no.~3, pp.~265--271, 1996.

\bibitem{ziegler}
A.~Ziegler, ``Word length in portuguese texts.,'' {\em Journal of Quantitative
  Linguistics}, vol.~5, no.~1-2, pp.~115--120, 1998.

\bibitem{mont2002}
M.~A. Montemurro and P.~A. Pury, ``Long-range fractal correlations in literary
  corpora,'' {\em Fractals}, vol.~10, pp.~451--461, 2002.

\bibitem{zanette05}
D.~Zanette and M.~Montemurro, ``Dynamics of text generation with realistic
  {Z}ipf's distribution,'' {\em Journal of Quantitative Linguistics}, vol.~12,
  no.~1, pp.~29--40, 2005.

\bibitem{mont2011}
M.~A. Montemurro and D.~H. Zanette, ``{Universal Entropy of Word Ordering
  Across Linguistic Families},'' {\em PLoS ONE}, vol.~6, no.~5, p.~e19875,
  2011.

\bibitem{ausloos2007}
R.~Lambiotte, M.~Ausloos, and M.~Thelwall, ``Word statistics in {B}logs and
  {RSS} feeds: Towards empirical universal evidence,'' {\em Journal of
  Informetrics}, vol.~1, no.~4, pp.~277--286, 2007.

\bibitem{ausloos2008}
M.~Ausloos, ``Equilibrium and dynamic methods when comparing an english text
  and its esperanto translation,'' {\em Physica A: Statistical Mechanics and
  its Applications}, vol.~387, no.~25, pp.~6411--6420, 2008.

\bibitem{ausloos2012}
M.~Ausloos, ``Generalized {H}urst exponent and multifractal function of
  original and translated texts mapped into frequency and length time series,''
  {\em Phys. Rev. E}, vol.~86, p.~031108, 2012.

\bibitem{sahin}
G.~Şahin, M.~Erentürk, and A.~Hacinliyan, ``Detrended fluctuation analysis in
  natural languages using non-corpus parametrization,'' {\em Chaos, Solitons
  and Fractals}, vol.~41, no.~1, pp.~198--205, 2009.

\bibitem{koehn2005}
P.~Koehn, ``{Europarl: A Parallel Corpus for Statistical Machine
  Translation},'' in {\em {Conference Proceedings: {T}he tenth Machine
  Translation Summit}}, (Phuket, Thailand), pp.~79--86, 2005.

\bibitem{kalimeri2012}
M.~Kalimeri, V.~Constantoudis, C.~Papadimitriou, K.~Karamanos, F.~Diakonos, and
  H.~Papageorgiou, ``Entropy analysis of word-length series of natural language
  texts: Effects of text language and genre,'' {\em International Journal of
  Bifurcation and Chaos}, vol.~22, no.~9, 2012.

\bibitem{nicolis1994}
G.~Nicolis and P.~Gaspard, ``Toward a probabilistic approach to complex
  systems,'' {\em Chaos, Solitons \& Fractals}, vol.~4, no.~1, pp.~41--57,
  1994.

\bibitem{karamanos1999}
K.~Karamanos and G.~Nicolis, ``Symbolic dynamics and entropy analysis of
  {F}eigenbaum limit sets,'' {\em Chaos, Solitons, \& Fractals}, vol.~10,
  no.~7, pp.~1135--1150, 1999.

\bibitem{nicolis1991}
W.~Ebeling and G.~Nicolis, ``Entropy of symbolic sequences: The role of
  correlations,'' {\em EPL (Europhysics Letters)}, vol.~14, no.~3, p.~191,
  1991.

\bibitem{nicolis2005}
J.~S. Nicolis, ``Super-selection rules modulating complexity: an overview,''
  {\em Chaos, Solitons \& Fractals}, vol.~24, no.~5, pp.~1159--1163, 2005.

\bibitem{nicolis2007}
J.~S. Nicolis, ``The role of chaos in cognition and music - super selection
  rules moderating complexity - a research program,'' {\em Chaos, Solitons \&
  Fractals}, vol.~33, no.~4, pp.~1093--1094, 2007.

\bibitem{karamanos2005}
J.~Borwein and K.~Karamanos, ``Algebraic dynamics of certain gamma function
  values,'' in {\em Generalized Convexity, Generalized Monotonicity and
  Applications} (A.~Eberhard, N.~Hadjisavvas, and D.~Luc, eds.), vol.~77 of
  {\em Nonconvex Optimization and Its Applications}, pp.~3--21, Springer US,
  2005.

\bibitem{borwein2008}
J.~M. Borwein and D.~H. Bailey, {\em {M}athematics by experiment}.
\newblock {A} {K} {P}eters {L}td., {S}econd~ed., 2008.

\bibitem{montemurro2013}
M.~A. Montemurro, ``Quantifying the information in the long-range order of
  words: Semantic structures and universal linguistic constraints,'' {\em
  Cortex}, 2013, http://dx.doi.org/10.1016/j.cortex.2013.08.008.

\bibitem{dunnevolved2011}
M.~Dunn, S.~Greenhill, S.~Levinson, and R.~Gray, ``Evolved structure of
  language shows lineage-specific trends in word-order universals,'' {\em
  Nature}, vol.~473, no.~7345, pp.~79--82, 2011.

\end{thebibliography}

\newpage

\appendix
\section*{Supporting Information}
\begin{figure}[h]
\centerline{\includegraphics[width=.75\textwidth]{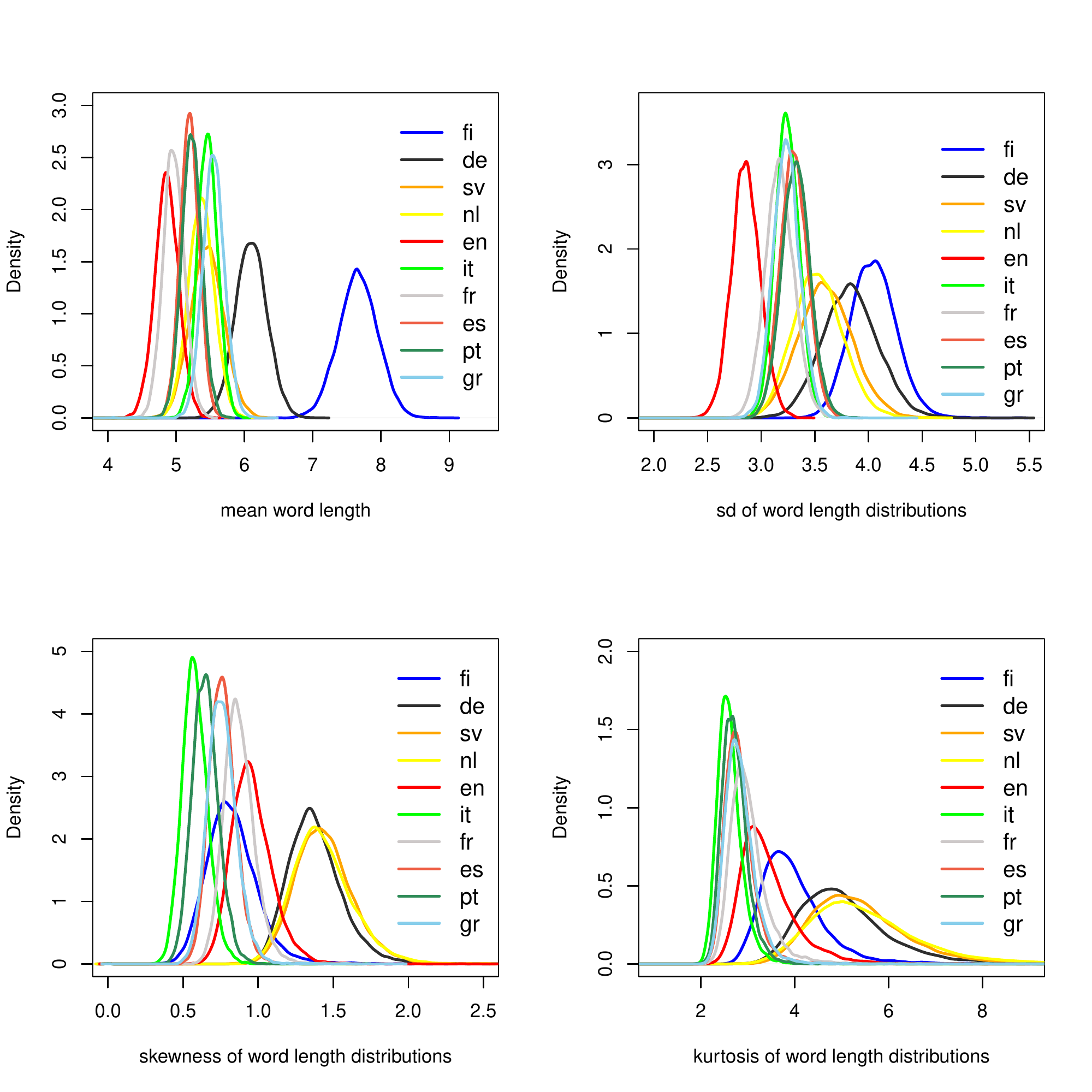}}
\caption{Figure S1: Distributions of the four moments, the mean value of which is given in Figure 3 of the main text, over all the different blocks of length 1000. To facilitate visualization, each curve is a Gaussian kernel density estimate of the corresponding histogram.}
\end{figure}

\begin{figure}[h!]
\centerline{\includegraphics[width=\textwidth]{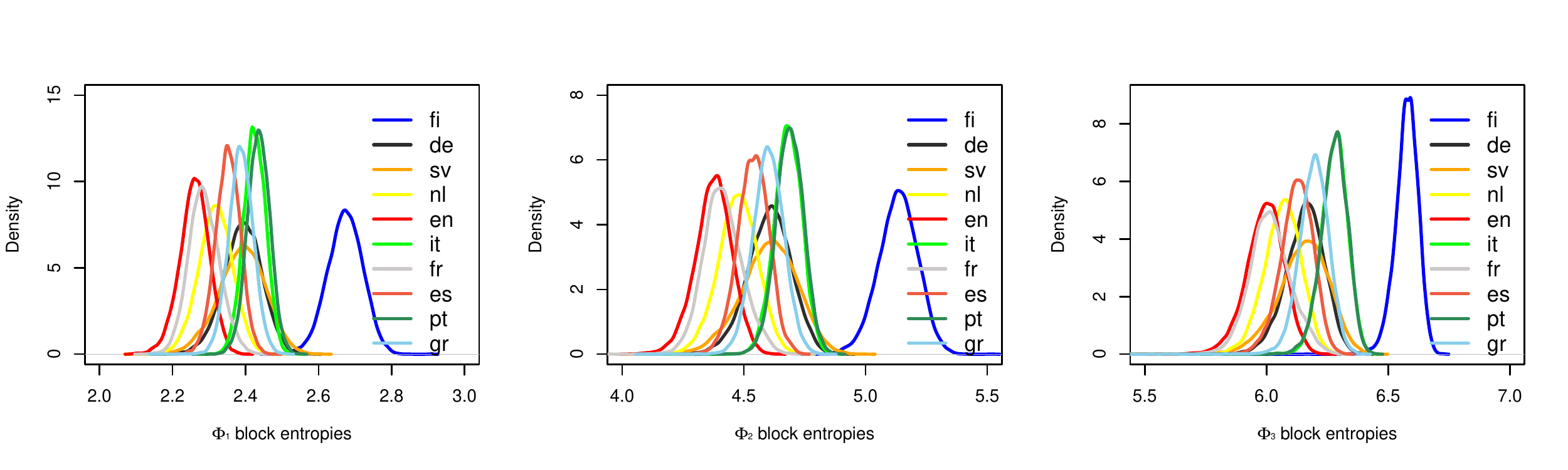}}
\caption{Figure S2: Distributions of the three $\Phi_i$ entropies, the mean value of which is given in Figures 5 and 6 of the main text, over all the different blocks of length 1000. To facilitate visualization, each curve is a Gaussian kernel density estimate of the corresponding histogram.}
\end{figure}

\end{document}